\pdfoutput=1

\documentclass[11pt]{article}

\usepackage[review]{ACL2023}

\usepackage{times}
\usepackage{latexsym}

\usepackage[T1]{fontenc}

\usepackage[utf8]{inputenc}

\usepackage{microtype}

\usepackage{inconsolata}

\usepackage{booktabs}
\usepackage{graphicx}
\usepackage{subcaption}
\usepackage{amsmath}
\usepackage{geometry}
\usepackage[linguistics]{forest}
\usepackage{natbib}
\usepackage{hyperref}
\usepackage{afterpage}
\usepackage{tabularx}
\usepackage{multirow}
\newcolumntype{L}[1]{>{\raggedright\arraybackslash}p{#1}}

\title{  
Diagnosing Medical Datasets with Training Dynamics\\
}
\author{Laura Wenderoth \\
  Newnham College \\
  University of Cambridge\\
  {\tt lw754 (at) cam.ac.uk} }

\begin{document}
\maketitle
\begin{abstract}
This study explores the potential of using training dynamics as an automated alternative to human annotation for evaluating the quality of training data. The framework used is \textit{Data Maps}, which classifies data points into categories such as easy-to-learn, hard-to-learn, and ambiguous \cite{swayamdipta2020dataset}. \citet{swayamdipta2020dataset} highlight that difficult-to-learn examples often contain errors, and ambiguous cases have a significant impact on model training. To confirm the reliability of these findings, we replicated the experiments using a challenging dataset, with a focus on medical question answering. In addition to text comprehension, this field requires the acquisition of detailed medical knowledge, which further complicates the task. A comprehensive evaluation was conducted to assess the feasibility and transferability of the \textit{Data Maps} framework to the medical domain. The evaluation indicates that the framework is not suitable for addressing the unique challenges of datasets in answering medical questions. 
\end{abstract}

\section{Introduction}
Significant advancements have been achieved in the realm of natural language processing (NLP), and language generation models have become essential tools for various daily tasks \cite{George_George_2023,Chang_Wang_Wang_Wu_Yang_Zhu_Chen_Yi_Wang_Wang_et,Huo_Arabzadeh_Clarke_2023}. Nevertheless, a persistent challenge is that these models often generate hallucinated or incorrect responses \cite{Rawte_Sheth_Das_2023,Dhuliawala_Komeili_Xu_Raileanu_Li_Celikyilmaz_Weston_2023,Huang_Yu_Ma_Zhong_Feng_Wang_Chen_Peng_Feng_Qin_et,Huo_Arabzadeh_Clarke_2023}. In fields such as medicine, where precision is paramount, there is little tolerance for inaccuracies in language model (LLM) results. Therefore, there are ongoing efforts to fine-tune LLMs for medical datasets and enhance them with domain-specific expertise \cite{Lee_Yoon_Kim_Kim_Kim_So_Kang_2020, Singhal_Azizi_Tu_Mahdavi_Wei_Chung_Scales_Tanwani_Cole-Lewis_Pfohl_et,Singhal_Tu_Gottweis_Sayres_Wulczyn_Hou_Clark_Pfohl_Cole-Lewis_Neal_et, Luo_Zhang_Fan_Yang_Wu_Qiao_Nie_2023}. 

The accuracy of these models is often assessed using existing multiple choice datasets such as USMLE MedQA \cite{medqa-dataset}. However, the accuracy of these models is heavily reliant on the quality of the training data \cite{Liang_Tadesse_Ho_FeiFei_Zaharia_Zhang_Zou_2022,Bernhardt_Castro_Tanno_Schwaighofer_Tezcan_Monteiro_Bannur_Lungren_Nori_Glocker_et,Chklovski_Parks_Woodcroft_Tyson_2023,johnson2023assessing}. Generating and evaluating high-quality medical training data is a challenging task, as it must be done by experts, which is expensive and time-consuming. Several researchers have investigated training dynamics to evaluate data points during the training process \cite{Xing_Arpit_Tsirigotis_Bengio_2018, toneva2018empirical, Bras_Swayamdipta_Bhagavatula_Zellers_Peters_Sabharwal_Choi_2020,swayamdipta2020dataset}. \citet{swayamdipta2020dataset} propose a promising solution to evaluate datasets based on training dynamics using a framework called \textit{Data Maps}. This approach divides the training instances in a dataset into easy-to-learn, hard-to-learn and ambiguous classes.
It was discovered that training on only 33\% of the dataset, but with examples solely from the ambiguous class, produces comparable results on the out-of-distribution (OOD) dataset as training on the entire dataset.  Additionally, it has been demonstrated that data points that are hard-to-learn have a higher prevalence of mislabelled instances than their easy-to-learn counterparts. This finding suggests that it is feasible to identify valuable and accurate instances of adversarial datasets. Once created, hard-to-learn examples can be filtered out and particularly ambiguous examples can be used for further training to reduce the risk of error and increase generalisability.

This approach is particularly valuable in the medical field, where dataset creation requires expert collaboration, making it expensive and labour-intensive.  The \textit{Data Maps} framework can be a cost-effective tool for quality assessment in medical datasets, contributing to a more robust and reliable application of LLMs in medical contexts.
Therefore, this paper makes the following contributions:
\begin{itemize}
    \item Replication: Repeating the experiment on a new dataset to demonstrate overall effectiveness.
    \item Feasability Evaluation: Assessing the practicability of the \textit{Data Maps} framework on datasets in general. This entails a meticulous examination of the method's adaptability and efficacy.
    \item Transferability to  Medical Domain: Analyse method's inherent strengths and weaknesses, particularly within the intricate domain of medical question answering.
\end{itemize}

\section{Data} 
\begin{figure*}[ht!]  
  \centering
  \includegraphics[width=0.8\textwidth]{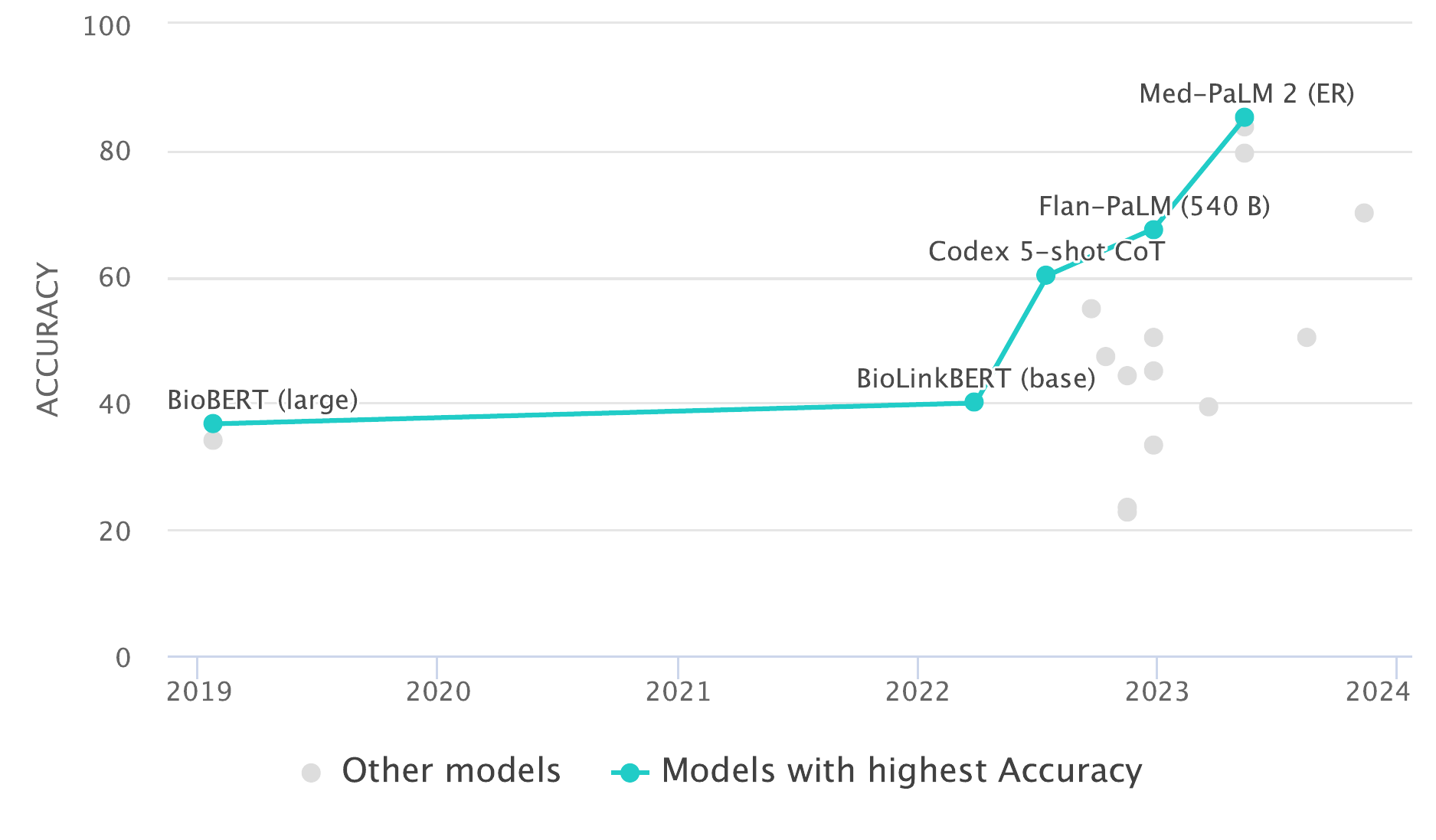}  
  \caption{The ranking displays how well various models performed in answering medical questions, as evaluated against the MedQA benchmark. The ranking is determined by accuracy and offers a thorough assessment of each model's effectiveness in answering medical questions. Chart borrowed from \citet{Medqabenchmark}.}
  \label{fig:medqa-benchmark}
\end{figure*}

This study uses two multiple-choice datasets. The primary dataset used for training is the MedQA dataset, which contains a comprehensive set of medical knowledge questions. To evaluate the  out-of-distribution (OOD) model's performance, we also incorporated questions from the Applied Knowledge Test of the Royal College of General Practitioners  general practitioners (GP) examination in the UK. This dataset is referred to as GP-UK. 

\paragraph{MedQA} is derived from the United States Medical Licence Exams (USMLE), providing a solid foundation based on professional medical exams \cite{medqa-dataset}. The dataset covers 2.4k healthcare topics and spans 21 different medical subjects. The dataset contains historical exam questions from official websites such as AIIMS and NEET PG, covering the period from 1991 to the present. An exam question comprises a single query with four potential solutions, and the most correct solution has to be found. It is divided into training (182,822 samples), validation (4,183 samples) and test data (6,150 samples).

The dataset has been used for benchmarking since 2019. Figure~\ref{fig:medqa-benchmark} shows the ranking created by \citet{Medqabenchmark}. The
Med-PaLM~2 model ranks highest with an exceptional accuracy of 85.4\%, followed closely by alternative configurations of the same model labelled CoT~+~SC (83.7\%) and 5-shot (79.7\%). Passing these exams typically requires a score of 60\%. 
However, it is important to note that prior to 2022, language models such as BioLinkBERT base and large, and GPT-Neo (2.7~B) only achieved accuracies of around or less than 40\%. This demonstrates the historical difficulty in obtaining accurate answers to medical questions using language models.

\paragraph{GP-UK} comprises sample questions from two Applied Knowledge Tests created by the \citet{akt-example-questions,mrcgp-akt}. Questions accompanied by pictures were excluded, resulting in a total of 59 questions. Furthermore, the answer options for the questions in the GP-UK dataset were restricted to four, which aligns with the standard multiple-choice format of medical exams in the US. This curated dataset serves as an OOD dataset for our investigation of the effectiveness and generalisability of the proposed methodology in the context of answering medical questions.

\section{Mapping Datasets with Training Dynamics}
In the field of dataset evaluation, several approaches have been explored, each offering unique perspectives on evaluating the quality and relevance of data. 
One possibility is to look at various parameters that change during training. For example, \citet{toneva2018empirical} analyse the training dynamics, focusing in particular on instances that are often misclassified in later training epochs, even though they were previously classified correctly.

We are examining a comparable technique developed by \citet{swayamdipta2020dataset}, known as \textit{Data~Maps}.
The main aim of \textit{Data Maps} is to visually represent how a model interacts with a dataset over time, allowing for a detailed understanding of the contributions made by individual instances to the model's learning process.
To achieve this, three measures are used: confidence, variability, and correctness.  The metrics are calculated by making a full pass over the training set after each epoch, without gradient updates, and saving the probabilities for the true label of each sample.
These measures capture key aspects of the model's behaviour during training, revealing its level of certainty, prediction consistency, and classification accuracy across different instances.

\paragraph{Confidence $\hat{\mu}_i$} quantifies the model's reliability in assigning the correct label to an observation. It is formally expressed as the average model probability assigned to the true label \(y^*_i\) over the entire training process of \(E\) epochs, as denoted by the Equation \ref{eq:confidence}:
\begin{equation}
    \hat{\mu}_i = \frac{1}{E} \sum_{e=1}^{E} p_{\theta(e)}(y^*_i \,|\, x_i)
    \label{eq:confidence}
\end{equation}
where \( p_{\theta(e)} \) is the probability of the model with parameters \( \theta(e) \) at the end of the \( e^{th} \) epoch.

\paragraph{Variability $\hat{\sigma}_i$} measures the distribution of the model's label assignment probabilities $p_{\theta(e)}$ for a given instance $y^*_i$ across epochs $E$. The standard deviation, as shown in Equation \ref{eq:variability}, quantifies it:
\begin{equation}
    \hat{\sigma}_i = \sqrt{\frac{ \sum_{e=1}^{E} \left(p_{\theta(e)}(y^*_i \,|\, x_i) - \hat{\mu}_i\right)^2} {E}}
    \label{eq:variability}
\end{equation}
The variability value indicates the consistency of the model's predictions for a specific instance. Low variability suggests stable label assignment, while high variability suggests indecision during training.

\paragraph{Correctness}  is a metric that reflects how accurately the model labels observations over epochs. It has possible values of $1+E$ and indicates the accuracy of the model's predictions over the entire training period, normalised by the maximum number of correct classifications on a dataset.

\section{Experiments}
The methodology of the study involves training a RoBERTa-large model on the entire training dataset, selecting appropriate instances, and then retraining the model using only those instances. We utilize the pre-trained version of RoBERTa in English, trained using a masked language modelling objective, to finetune it on the medical question answering dataset. Additionally, we test this pre-trained version without fine-tuning of RoBERTa on both the MedQA and OOD dataset. We choose exclusivly the RoBERTa model because it was identified as optimal in the original paper. 
Each training phase consists of 20 epochs, with an early stopping after 10 epochs at constant validation accuracy and a batch size of 96 epochs. This is in line with the specifications of the original paper. The training process utilises an RTX A6000 GPU. 

First, the model was trained using the entire training dataset. The \textit{Data Maps} framework was applied to compute the three training dynamics. These were used to classify training examples as easy-to-learn, ambiguous, or hard-to-learn.
\paragraph{Easy-to-learn} examples have high confidence and low variability, with no set limit for either. These values depend on the training data. For a predefined proportion, data with the highest confidence and lowest variability is selected as easy-to-learn until the designated proportion of data is reached.

\paragraph{Hard-to-learn} examples, on the other side, have low confidence and low variability. There are no limits for either, and it simply represents  a distribution from which a certain percentage is selected. 

\paragraph{Ambiguous} examples are the only one with a high variability. This means that the probabilities of the true classes fluctuate frequently during training.\vspace{10pt}
\\
After training the model on the entire dataset and calculating the training dynamics, two additional experiments are conducted using only 33\% of the training data. In the first run, 33\% of the training data is randomly sampled. In the second experiment, the 33\% with the highest variability is selected - the ambiguous examples.
Subsequently, all three models are evaluated on the test dataset and the OOD dataset under the metric accuracy. The corresponing code has been published on github \href{https://github.com/LauraWenderoth/training-dynamics}{\texttt{LauraWenderoth/training-dynamics}} for the purpose of reproducibility.


\begin{table}[t!]
    \centering
    \caption{RoBERTa-large model performance on MedQA (ID) and out-of-distribution (OOD) generalization. This table presents the performance of the RoBERTa-large model on the MedQA dataset under different training scenarios. First, the open available pre-trained RoBERTa-large model \cite{robertalarge} was applied to both datasets. Than, the model was trained using a single seed on different proportions of the dataset, including 100\% and two subsets of 33\%. The 33\% subsets were created by randomly selecting instances and targeting those with the highest ambiguity. The best outcomes are highlighted in bold.}
    \begin{tabularx}{\linewidth}{@{}ll>{\centering\arraybackslash}X>{\centering\arraybackslash}X@{}}
        \toprule
        & &MedQA Test (ID) & GP-UK (OOD) \\
        \midrule
        \multicolumn{2}{l}{pretrained}  & 31.69& 25.42 \\
        \midrule
        \multicolumn{2}{l}{100\% train}  & \textbf{36.07}& \textbf{30.50} \\
        \addlinespace[0.5em] 
        \multirow{2}{*}{\rotatebox{90}{33\%}} & random & 33.38& 20.34\\
         & ambiguous & 33.02 & 22.03\\
        \addlinespace[0.5em] 
              \bottomrule
    \end{tabularx}
    \label{table:results}
\end{table}

\section{Results}
\begin{figure*}[ht!]  
  \centering
  \includegraphics[width=\textwidth]{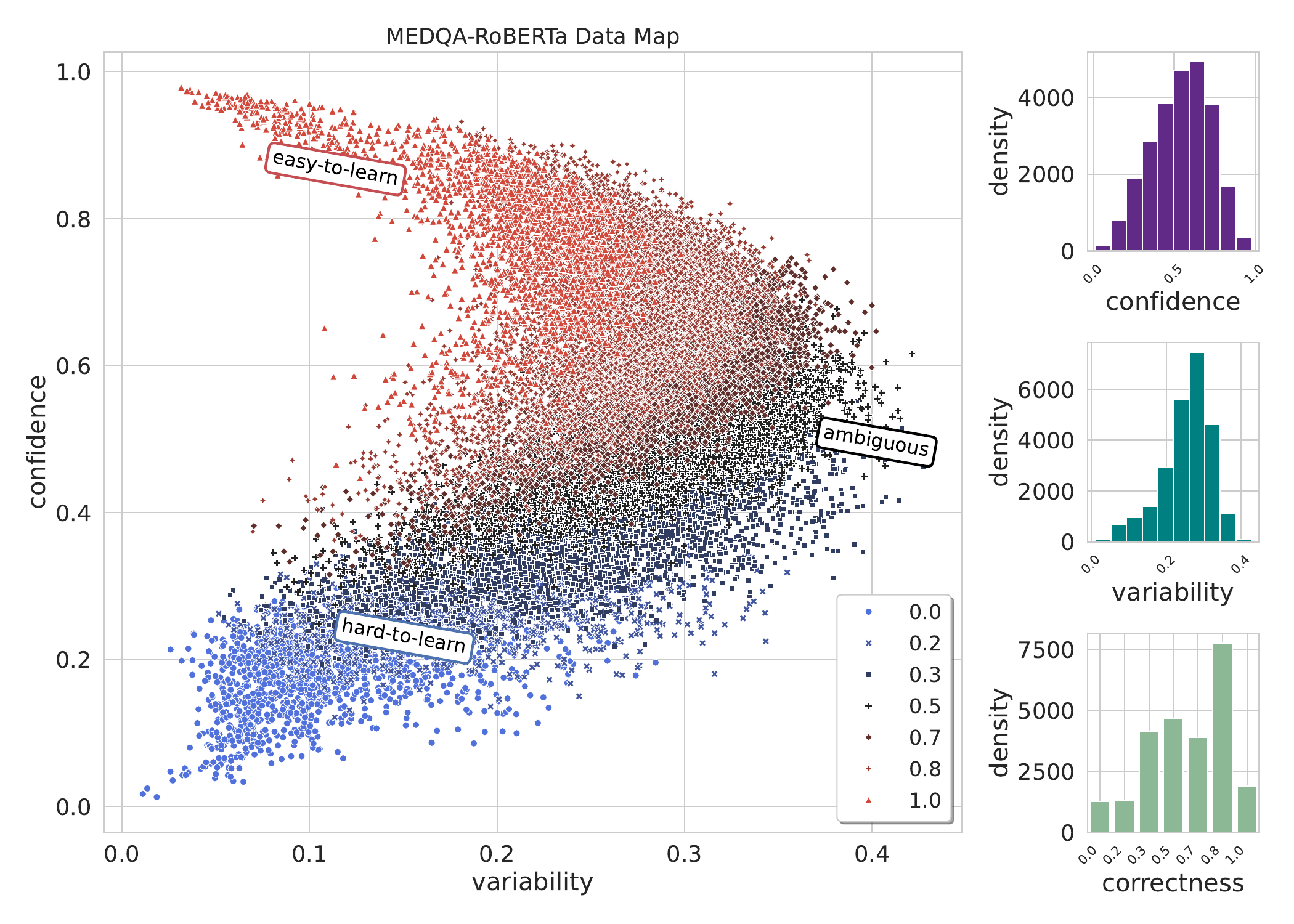}  
  \caption{\textit{Data Maps} analysis on the MedQA dataset using the performance of the ROBERTA-large classifier over 20 epochs with 182,822 training samples. Only 25,000 samples are displayed for clarity. The x-axis indicates variability, ranging from low to high, while the y-axis represents the confidence levels of the classifier. The visualisation uses different colours and shapes to indicate correctness. Red triangles represent easy-to-learn examples with low variability and high confidence, blue circles represent hard-to-learn instances with low variability and low confidence, and black pluses represent ambiguous cases with high variability. This intuitive representation provides a comprehensive overview of the dataset with respect to the classifier. }
  \label{fig:datamet-main-roberta}
\end{figure*}

The results are divided into three different subsections: replication, feasibility, and transferability to the medical domain. \textit{Replication} examines the performance of the training dynamics calculated using the \textit{Data Maps} framework on the MedQA dataset, comparing the results with those of the original work. The \textit{feasibility} section assesses the technical implementation by evaluating the ease of replication and efficiency of creating \textit{Data Maps}. In the final section, \textit{transferability to the medical domain}, we evaluate the potential value and applicability of these training dynamics in a medical context. 

\subsection{Replication}
The replication section is divided into \textit{data selection} and \textit{detection of mislabelled instances}, two tasks from the original paper replicated on the MedQA dataset. Data selection aims to filter instances that are valuable for training to enhance generalizability. The detection of mislabelled instances is based on findings by \citet{swayamdipta2020dataset}, which suggest a higher prevalence of mislabelled samples in the hard-to-learn category compared to the easy-to-learn category.

\paragraph{Data Selection} Table \ref{table:results}  provides an overview of the model's performance in various training scenarios. The best results are achieved when training on the full MedQA dataset, which shows superior performance on both the MedQA testset and the OOD dataset. When using the 33\% training subset, random selection appears to perform better on the ID dataset, although this difference could be due to chance with only one trained seed. Therefore it is very likely that random and ambiguous selection perform similarly well on 33\% of the ID dataset, in contrast to the original paper which found a significant difference between these two selections. It is worth noting that the ambiguous selection of 33\% in the original paper approaches the same training accuracy as the the 100\%. However, a clear improvement can be seen compared to the pre-trained model.

The random selection in the OOD dataset performed worse that the ambiguous selection just by 2 percentage points which correlates to one misclassified instance. From this it can be deduced that both selections perform equally well on the OOD dataset, given the restriction to one seed and the small sample size. Notably, both selections perform worse than chance and consequently worse than the pretrained model, which performs as well as chance. This discrepancy can be attributed to the uneven distribution of response options in the OOD dataset. In contrast to the original work, the model trained on the entire dataset generalizes better than the model trained on the ambiguous selection. But this is one of the significant contributions of \textit{Data~Maps}. However, this result could not be replicated for MedQA.

\paragraph{Detection of Mislabelled Instances} When investigating mislabelled examples, we manually checked the 100 most hard-to-learn examples. We comprehensively reviewed each example to ensure it was correct and matched the provided label.  We did not identify any incorrectly labelled examples during this rigorous review. As the questions in our dataset are sourced from official exam questions, the probability of common errors is low. This is in contrast to other open-source datasets, such as SNLI \cite{bowman2015large}, which were used in the original paper \cite{swayamdipta2020dataset}. The absence of mislabelled examples among the hard-to-learn instances highlights the dataset's reliability and robustness.

In summary, we were unable to replicate the results presented in the original paper. We found no significant difference in the performance of models trained on randomly selected or ambiguous subsets of the dataset. Additionally, our investigation of potentially mislabelled examples revealed no inaccuracies. 

\begin{figure}[t!]  
  \centering
  \includegraphics[width=\columnwidth]{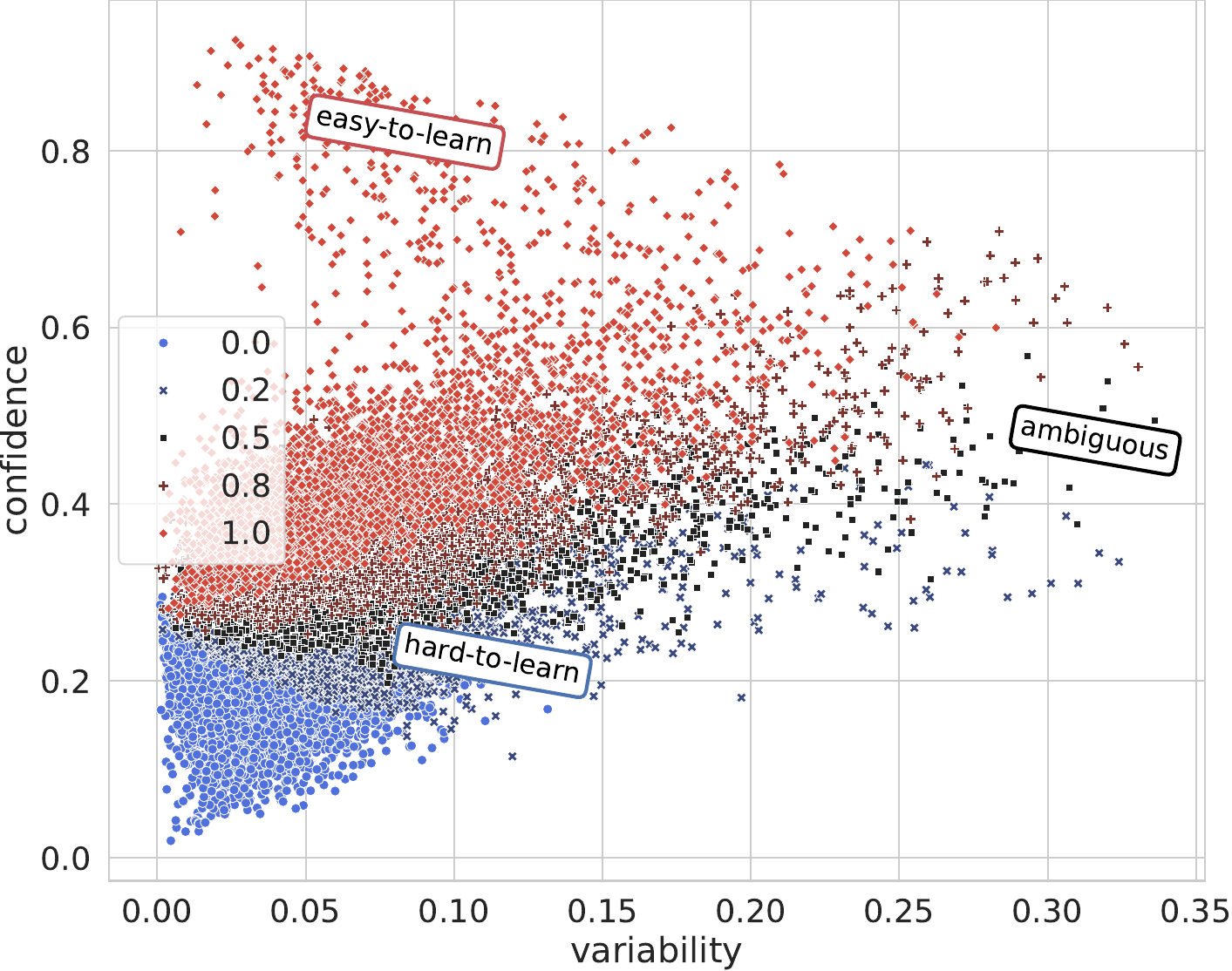}  
  \caption{\textit{Data Maps} analysis on the MedQA dataset using the performance of the ROBERTA-large classifier like in Figure \ref{fig:datamet-main-roberta}. The training dynamics were only calculated for five epochs instead of the full 20. This causes the data points to be closer together, making them harder to distinguish.}
  \label{fig:6epoch}
\end{figure}

\subsection{Feasibility}
The implementation of the provided code was technically challenging due to its lack of modularity and inadequate documentation. Critical information, such as requirements and Python version compatibility, was absent, making it difficult to adapt the codebase. The code's inflexibility compounded the difficulty of applying it to new datasets, requiring significant effort. Apart from creating \textit{Data Maps} plots, which were relatively straightforward once the data had been computed, other aspects of the codebase were less reproducible. 

In terms of efficiency, training the model for a single epoch on an RTX A6000 GPU with a batch size of 96 took an average of 47 minutes and 37 seconds. The training was conducted over 20 epochs, resulting in a training time of 16 hours without the calculation of training dynamics. Overall, the calculation took approximately 18 hours. The results are illustrated in Figure \ref{fig:datamet-main-roberta}. These findings illuminate the computational demands of the training process to be able to replicate the training dynamics using the \textit{Data Maps} framework.
\begin{figure}[t!]  
  \centering
  \includegraphics[width=\columnwidth]{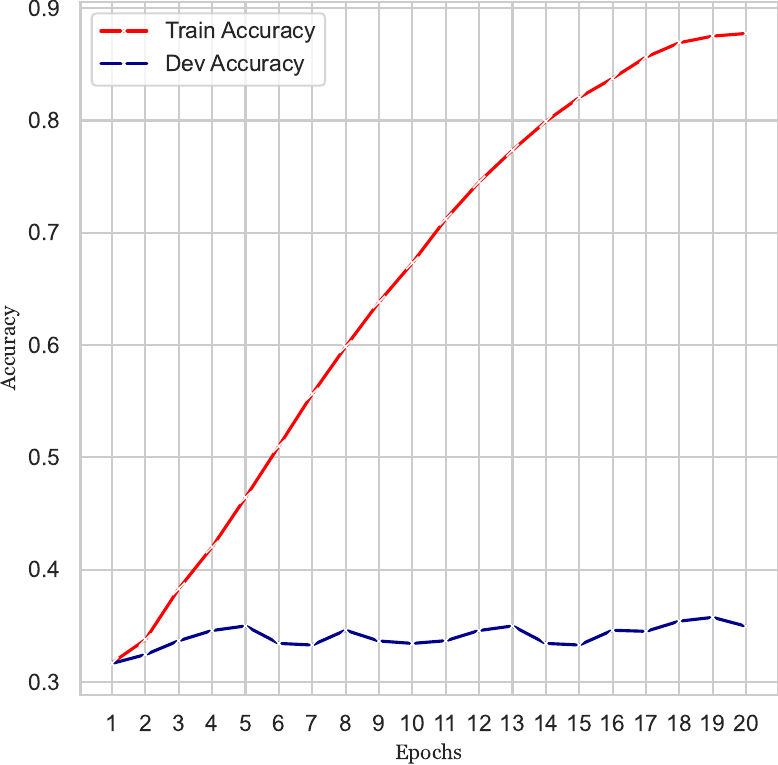}  
  \caption{RoBERTa-large model performance over training epochs. The plot indicates that the model is overfitting on the training data, resulting in no improvement on the validation data. The training accuracy increased rapidly during the training process, reaching a maximum of 87.7\% in the 20th epoch. In contrast, the validation accuracy initially increased up to epoch 5, followed by fluctuation, and reached a maximum of 35.74\% at epoch 19.}
  \label{fig:trainperformance}
\end{figure}


\subsection{Transferability to Medical Domain}
Training a model to answer medical questions is a challenging task, as demonstrated by the performance chart on the MedQA dataset in Figure \ref{fig:medqa-benchmark}. BERT-based models have difficulty achieving an accuracy of 40\%. Therefore, our achievement of 36\% accuracy on the testset is commendable. However, it also indicates that the model must heavily overfit on the training data in order to achieve separation between data points calculated using traning dynamics. Figure \ref{fig:trainperformance} shows the training and validation accuracy over the epochs. The training and validation processes began with an initial accuracy of 31.7\%. The accuracy quickly increased during training, reaching a peak of 87.7\% by Epoch 20. In contrast, the validation accuracy initially increased up to Epoch 5, but then fluctuated with intermittent declines and recoveries. Finally, the validation accuracy stabilised and experienced a slight increase, reaching its maximum of 35.739\% in epoch 19. Most importantly Figure \ref{fig:trainperformance} illustrates that the model overfitts on the training data without any improvement on the evaluation accurary. 

However, not performing this step results in crowded scores, as ilustrated in Figure \ref{fig:6epoch}. In this case, the model was trained for only 5 epochs before calculating the training dynamics like it was done in the original paper. Medical datasets require significantly longer training than standard NLP datasets like SNLI. 
To address this issue, more complex models can be used to calculate training dynamics. LLMs like Med-PaLM2 achieve an accuracy rate of over 80\%, which is comparable to the standard NLP datasets used in the original paper. However, these models require even longer training times, which makes it unrealistic to calculate training dynamics.

This phenomenon illustrates the challenge of transferring knowledge to the medical field. Medical questions require not only lexical understanding but also domain-specific knowledge. The results highlight that the medical dataset is too complex, making it impossible to calculate the meaningful training dynamics. It appears that the significance of the training dynamics is only meaningful if the model learns from the training data and and does not simply overfit to it. Therefore, the transferability of the \textit{Data Maps} framework to medical data has not been successful.



\section{Conclusion}
In conclusion, \textit{Data Maps} provide an automated method for visualising and diagnosing large datasets through training dynamics. Our attempt to replicate the original paper's results was challenging, as we found no significant disparities in model performance between randomly selected and ambiguous subsets of the dataset. Furthermore, our examination of potentially mislabelled examples revealed no inaccuracies.

Feasibility assessments highlighted the resource-intensive nature of the approach. For example, when calculating training dynamics, which includes training on a 50 GB GPU, using the RoBERTa model required approximately 18 hours, resulting in significant computational and energy costs.

Our findings suggest limited success in transferring to the medical domain, especially for smaller models. \textit{Data Maps} classes, such as hard-to-learn, easy-to-learn, and ambiguous, could only be computed through overfitting. The ambiguous subset selected for the study did not perform better than randomly selected subset, which challenges the original paper's results. Large language models (LLMs) achieved similar accuracies on medical datasets to RoBERTa on the datasets used in the original paper. However, applying \textit{Data Maps} to LLMs and retraining appeared to be cost prohibitive.

Future research could explore the potential of using LLMs' hallucination capabilities to calculate the training dynamics confidence, variablity and correctness during inference. Developing a more cost-effective variant of \textit{Data Maps} for LLMs remains a challenge, leaving room for further exploration in subsequent studies.
\bibliography{project}
\bibliographystyle{acl_natbib}

\end{document}